\documentclass{article}

\usepackage[preprint]{neurips_2024}

\usepackage[utf8]{inputenc} % allow utf-8 input
\usepackage[T1]{fontenc}    % use 8-bit T1 fonts
\usepackage{hyperref}       % hyperlinks
\usepackage{url}            % simple URL typesetting
\usepackage{booktabs}       % professional-quality tables
\usepackage{amsfonts}       % blackboard math symbols
\usepackage{nicefrac}       % compact symbols for 1/2, etc.
\usepackage{microtype}      % microtypography
\usepackage{xcolor}         % colors

\usepackage{amsmath}
\usepackage{graphicx}  % Required for figures
\usepackage{subfigure} % Required for subfigures
\usepackage{pgfplots}  % Required for TikZ plots
\usepackage{pgfplotstable}  % May be needed for some pgfplots functionalities

\title{HNCSE: Advancing Sentence Embeddings via Hybrid Contrastive Learning with Hard Negatives}

\author{%
  Wenxiao Liu \\
  College of Cyberspace Security \\
  Jinan University \\
  \texttt{lwx\_mailbk@163.com} \\
  \And
  Zihong Yang \\
  College of Cyberspace Security \\
  Jinan University \\
  \And
  Chaozhuo Li \\
  School of Cyberspace Security \\
  Beijing University of Posts and Telecommunications \\
  \AND
  Zijin Hong \\
  Birmingham College \\
  Jinan University \\
  \And
  Jianfeng Ma \\
  College of Cyberspace Security \\
  Jinan University \\
  \And
  Zhiquan Liu \\
  College of Cyberspace Security \\
  Jinan University \\
  \AND
  Litian Zhang \\
  Beihang University \\
  \And
  Feiran Huang\thanks{Corresponding author.} \\
  College of Cyberspace Security \\
  Jinan University \\
}

\begin{document}

\maketitle

\begin{abstract}
Unsupervised sentence representation learning remains a critical challenge in modern natural language processing (NLP) research. Recently, contrastive learning techniques have achieved significant success in addressing this issue by effectively capturing textual semantics. Many such approaches prioritize the optimization using negative samples. In fields such as computer vision, hard negative samples (samples that are close to the decision boundary and thus more difficult to distinguish) have been shown to enhance representation learning. However, adapting hard negatives to contrastive sentence learning is complex due to the intricate syntactic and semantic details of text. To address this problem, we propose HNCSE, a novel contrastive learning framework that extends the leading SimCSE approach. The hallmark of HNCSE is its innovative use of hard negative samples to enhance the learning of both positive and negative samples, thereby achieving a deeper semantic understanding. Empirical tests on semantic textual similarity and transfer task datasets validate the superiority of HNCSE.
\end{abstract}

\section{Introduction}

Sentence representation learning (SRL), or sentence embedding, is a crucial subfield of natural language processing (NLP) that involves encoding sentences into low-dimensional vectors to capture their semantic content. 
The essence of learning sentence representations is to grasp the full linguistic context, transcending mere word meanings.
Sentence representation learning contributes to facilitating a myriad of applications, including information retrieval \cite{ir}, conversational AI \cite{cai}, and machine translation systems \cite{mts}.

Conventional endeavors generally follow the supervised learning paradigm \cite{jordan2013forward}. However, such supervised learning methods suffer from the substantial cost associated with gathering sufficient training data. 
Recently, unsupervised sentence learning aims to enhance efficiency and cost-effectiveness in training without sacrificing the accuracy or efficacy of the resulting sentence representations.
Unsupervised sentence representation learning refers to training models to understand the semantic content of sentences without explicit human-labeled guidance, often using methods like auto-encoding, predicting surrounding context, or self-supervision techniques such as contrastive learning \cite{SimCSE}. 
At present, the mainstream methods of unsupervised sentence representation learning follow the idea of contrastive learning, that is, to train the sentence representation model by maximizing the similarity between positive sample pairs and minimizing the similarity between negative sample pairs.

\begin{figure*}[htp]
  \centering
  \includegraphics[width=1\textwidth]{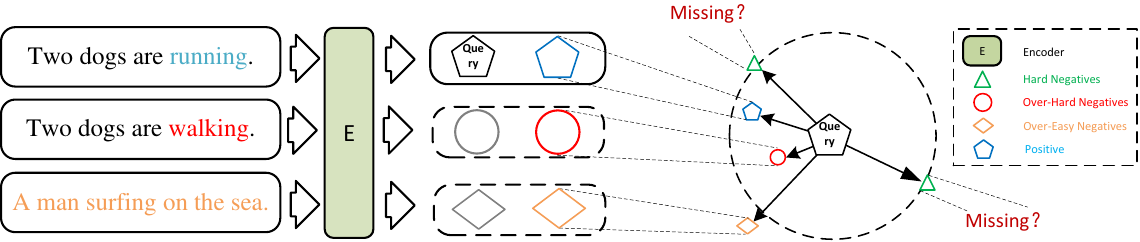}
  \caption{An example of a query with its positive, over-easy, over-hard and missing hard negative samples. The figure shows the situation that two too similar sentences may cause the loss of normal hard negative samples after passing through the encoder.
}
  \label{intro}
\end{figure*}

The core of contrastive learning-based SRL lies in how to construct effective \textbf{positive samples} and \textbf{negative samples}. 
Positive samples refer to the sentence pairs that are semantically similar or identical, while negative samples denote the sentence pairs that are semantically unrelated or distinct. 
Under unsupervised learning conditions, labeled information is not available. Therefore, we need to design some heuristic methods to generate correct positive and negative samples.
A popular strategy is to use the same sentence to generate positive samples through different data enhancement (such as random replacement, deletion, insertion of words, etc. \cite{diffcse}), and randomly select other sentences as the negative samples. 
However, such strategies are not a panacea to solve the unsupervised SRL task due to the following reasons. 
Firstly, the positive samples and negative samples are generated independently, without considering the semantic distance between them. 
Such a weak constraint may blur the boundary between positive and negative samples, leading to a dilemma such as the similarity between positive samples is lower than the one between negative samples \cite{SimCSE1}. 
Secondly, model performance is seriously affected by the number and distribution of positive and negative samples.  If the distributions of positive/negative samples are uneven, SRL models tend to learn biased or unstable sentence representations. 
Meanwhile,  the mainstream unsupervised SRL methods (e.g., SimCSE \cite{SimCSE}) adopt the method of in-batch negative sampling. Although in-batch negative sampling enjoys the advantage of computing resource efficiency, it still suffers from severe limitations such as relying on batch size, a limited number of negative samples, and diversity \cite{SimCSE2}.

This paper targets the acquisition of superior, correlated positive-negative sample pairs to enhance SRL. A major hurdle is managing confusing, over-hard, and over-easy negative samples effectively. Confusing samples, often a byproduct of data augmentation like in SimCSE, can lead to divergent embeddings for positive pairs, resulting in subpar learning samples. Over-hard negatives, due to their high similarity to the query, may cause learning distortion. Conversely, over-easy negatives, marked by their extreme dissimilarity, can lead to an unchallenging learning process, inhibiting the acquisition of meaningful representations. Figure \ref{intro} illustrates an example of a dense embedded distribution and the issue of missing hard negatives. Addressing these issues highlights the necessity for careful construction of positive/negative samples, particularly hard negatives that balance difficulty and simplicity. This balance is crucial for a well-tuned and effective learning path. However, identifying these challenging negatives remains a significant challenge due to computational and storage demands.

In this paper, we introduce an innovative unsupervised comparative learning framework that leverages hard negative sample mixing to mitigate the adverse effects of confusing samples and enrich the reliability of hard negatives within the same batch.
Contrasting prior research that solely targets negative samples, our approach underscores the significance of positive samples in unsupervised SRL, fostering a synergistic learning process between positive and negative samples.
Our framework not only harnesses hard negative information to enhance the quality of positive samples but also generates additional hard negatives through the mixing of existing samples. Furthermore, it efficiently expands the negative sample pool, facilitating convergence in alignment characteristics.
Extensive evaluations across various tasks have confirmed the superior performance of our proposal against SOTA benchmarks. In summary, our key contributions are as follows:

\begin{itemize}
    \item We propose the HNCSE framework, which is divided into HNCSE-PM and HNCSE-HNM, both extended based on SimCSE. HNCSE-PM constructs positive samples closer to the query through the hardest negative sample. HNCSE-HNM uses mixup on existing hard negative samples to obtain higher quality hard negative samples.
    \item The theoretical analysis of Hard Negative Mixing (HNM) has deeply elucidated its profound impact on the enhancement of sentence representation learning.
    \item Through extensive experiments, HNCSE has achieved promising improvements over SimCSE in terms of semantic textual similarity (STS) and transfer tasks (TR). Additionally, HNCSE demonstrates clear advantages over current popular large language models (LLMs) in STS tasks.
\end{itemize}

\section{Related Work}
In recent times, the academic community increasingly focuses on unsupervised sentence representation learning. Traditional approaches, such as those demonstrated by Arora et al.\cite{arora2017simple} and Ethayarajh \cite{ethayarajh2019contextual}, generate sentence embeddings through a weighted average of word embeddings. Kiros et al.\cite{kiros2015skip} introduce the SkipThought model, which applies the skip-gram model at the sentence level, using the encoded sentence to predict its adjacent sentences. Many researchers, like Qiao et al \cite{qiao2019understanding}, are adopting outputs from pre-trained language models like BERT for sentence embedding. However, Ethayarajh \cite{ethayarajh2019contextual} observes that directly using BERT for this task yields suboptimal results. Li et al \cite{BERT-Flow} note that BERT induces anisotropy in the sentence embedding space, adversely impacting the performance on Semantic Text Similarity (STS) tasks.

To address this, Li et al \cite{BERT-Flow} introduce BERT-flow, a method that utilizes a flow-based technique, as described by Dinh, Krueger, and Bengio \cite{dinh2014nice}. This technique transforms the anisotropic sentence embedding distribution into a more uniform, isotropic Gaussian distribution, aiming to improve performance on STS tasks. Concurrently, Su et al \cite{BERT-whitening} present the BERT-whitening approach. This method leverages traditional whitening techniques to produce a more consistent sentence embedding distribution. It not only matches the performance of BERT-flow but also reduces the dimensionality of the sentence embeddings, enhancing computational efficiency.

\section{Methodology}

The HNCSE model, as outlined in the literature, pioneers a novel training approach that leverages hard negative mixing to refine sentence embeddings. It rectifies misclassified positives by incorporating hard negative traits, thereby boosting the model's discriminative capacity. HNCSE then broadens the challenge set by augmenting existing hard negatives, fostering a more robust learning context that advances sentence representation learning. Following this, we will elucidate the foundation of our base model, SimCSE \cite{SimCSE}.

The core idea of unsupervised SimCSE is to use a contrastive learning objective to train sentence embeddings without any labeled data. The contrastive learning objective is based on the principle of maximizing the agreement between two different views of the same sentence, while minimizing the agreement between different sentences. The two views are obtained by applying dropout to the input sentence and encoding it with a pre-trained transformer model. The contrastive learning objective can be formulated as follows:
\begin{equation} 
\begin{aligned}
\mathcal{L}(\theta) = -\frac{1}{N}\sum_{i=1}^{N}\log\frac{\exp\left(\text{sim}\left(f_{\theta}(s_i), f_{\theta}(\tilde{s}_i)\right)/\tau\right)}{\sum_{j=1}^{N}\exp\left(\text{sim}\left(f_{\theta}(s_i), f_{\theta}(\tilde{s}_j)\right)/\tau\right)}
\end{aligned}
\end{equation}
where $f_{\theta}$ is the sentence encoder parameterized by $\theta$, $s_i$ and $\tilde{s}_i$ are two views of the $i$-th sentence, $\text{sim}$ is a cosine similarity function, $\tau$ is a temperature parameter, and $N$ is the batch size. This objective encourages the embeddings of the same sentence to be close to each other, while pushing away the embeddings of different sentences. This is a simple but effective method to learn sentence embeddings that can capture semantic similarity. 
Next we will detail each of the two ways to improve the unsupervised SimCSE.

\subsection{Positive Mixing}

\begin{figure}[t]
  \centering
  \includegraphics[width=0.7\textwidth]{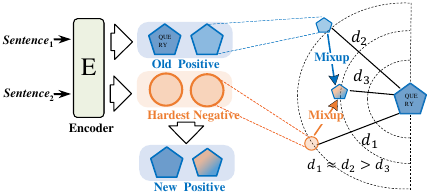}
  \caption{Schematic diagram of Positive Mixing.}
  \label{pos_fig}
\end{figure}

In text-based contrastive learning, "hard negatives" are exemplars that the model nearly misclassifies as positive during training. These negatives are erroneously deemed highly similar to positives.

Integrating hard negatives with positives offers unique advantages. It mirrors real-world scenarios where category distinctions can be ambiguous, training the model to navigate such complexities. This refines the model's class boundaries, enhancing its ability to discern between subtly distinct instances. It also mitigates overfitting by diverting the model's focus from only the clear and easy examples.

SimCSE's positive sampling involves feeding the same sentence into the model twice to yield two distinct eigenvectors as positives, facilitated by dropout layers. Despite dropout, semantically distinct sentences may still produce similar positive samples due to random masking.

For example, as shown in Figure \ref{pos_fig}, given two input sentences "Two dogs are running'' standing for $Sentence_1$ and "Two dogs are walking'' standing for $Sentence_2$, after dropout in SimCSE's unsupervised mode, two pairs of query and positive can be achieved, which are respectively shown as ($Q_1$, $P_1$) and ($Q_2$, $P_2$). The similarity between $Q_1$ and $P_2$ might be even larger that the one between $Q_1$ and $P_1$, which is obviously incorrect.

In order to shorten the distance between the query and the positive while increasing the distance between the positive and the negative, we compute the similarity between the query and the positive as well as the hardest negative, denoted as $d_1$ and $d_2$, respectively:

\begin{equation} 
s_2^{+} = \begin{cases} 
w_1s_1^{+}+w_2s_1^{-}, & 0<\frac{d_1-d_2}{d_1} \leq 0.1 \\ 
s_1^{+}, & \frac{d_1-d_2}{d_1}>0.1\\
\frac{d_1-d_2}{d_1} s_1^{+}+ \frac{d_1-d_2}{d_2 }s_1^{-}, &d_1 < d_2 
\end{cases}
\label{pos1}
\end{equation} 

In Equation (\ref{pos1}), $s_1^{+}$ denotes the positive example, while $s_1^{-}$ stands for the most similar negative instance. $s_2^{+}$ refers to the generated positive instance after the mixup of both positive and negative instances. $d_1$ signifies the similarity between the query and the positive instance, and $d_2$ refers to the similarity between the query and the hardest negative. $w_1$ and $w_2$ respectively represent the weights of $s_1^{+}$ and $s_1^{-}$.

\textbullet\enspace{\textbf{Contrastive Loss.}}  
InfoNCE is adopted as the contrastive loss, which represents the model capability to estimate the mutual information. Optimizing InfoNCE loss maximizes the mutual information between positive samples and minimizes the mutual information between negative samples as:

\begin{equation}
\mathcal{L}_i=-\frac{1}{B}\sum_{i=1}^{B}\log \frac{e^{\operatorname{sim}\left(s(i), s(i)^{\prime}\right) / \tau}}{\sum_{j=1}^{2B} e^{\operatorname{sim}\left(s(i), s(j)\right) / \tau}}
\end{equation}
where $\mathbf{s}(i)$ and $\mathbf{s}(i)^{\prime}$ are embeddings obtained after the same sentence is input to the encoder through different dropout masks, which are positive sample pairs; $\operatorname{sim}(\cdot, \cdot)$ is the cosine similarity function; $\tau$ is a temperature coefficient that regulates similarity scaling; $B$ is the number of sentences in a batch, and $2B$ is the number of vectors in a batch.

\subsection{Hard Negative Mixing}

\begin{figure*}[htp]
  \centering
  \includegraphics[width=0.8\textwidth]{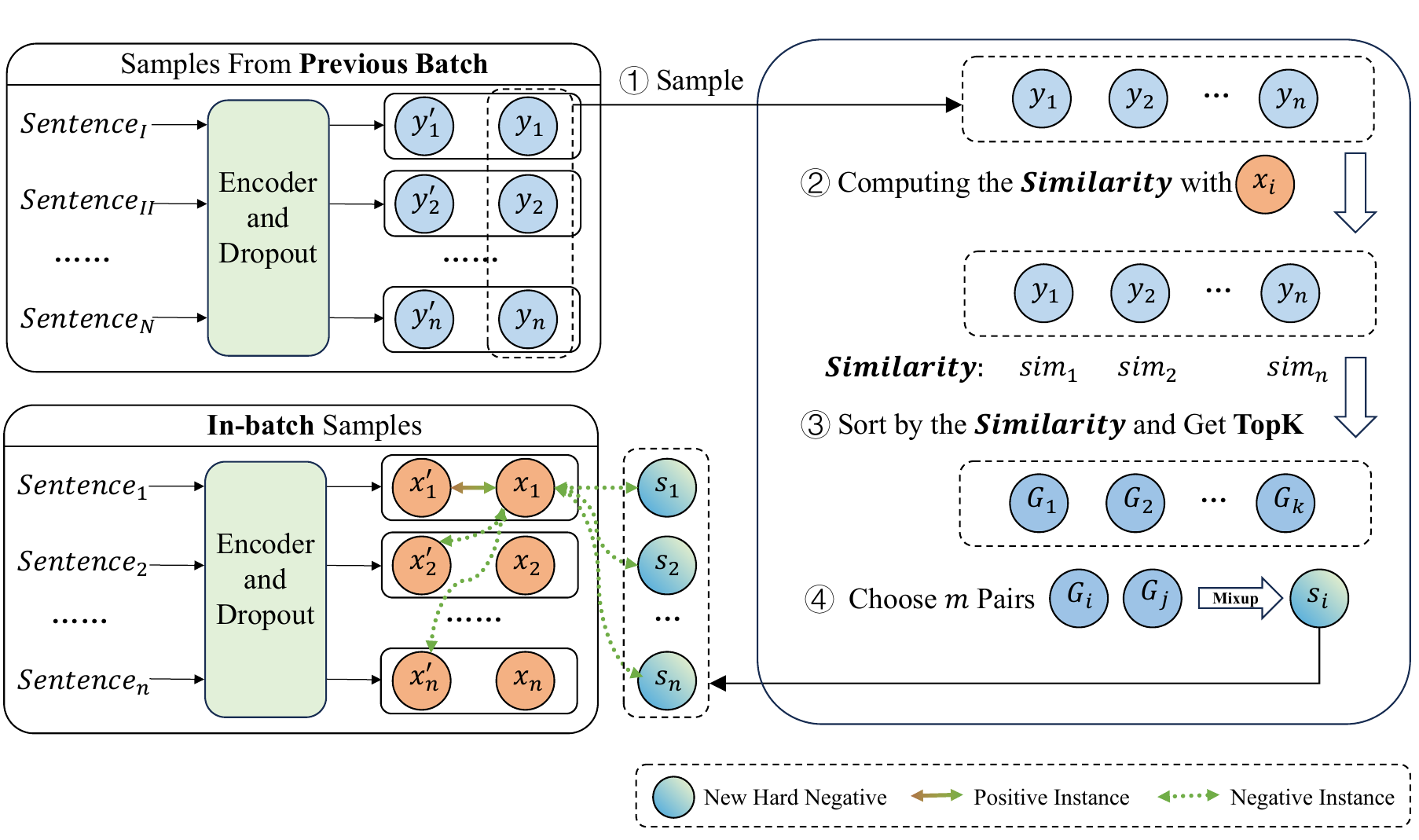}
  \caption{Schematic diagram of Hard Negative Mixing.}
  \label{fig:hnm}
\end{figure*}

Consider a scenario in which the embedding of a given query, denoted as $q$, and the embedding of a corresponding key in the matched pair, denoted as $k$, is contrasted with every feature $n$ in the bank of negatives ($N$).

The popular loss function for contrastive learning \cite{chen2020simple,he2020momentum} is:
\begin{equation}
\mathcal{L}{q,k,{N}} = - \log \frac{\exp (q^T k / \tau) }{ \exp (q^T k / \tau) + \sum{n \in N} \exp (q^T n / \tau)}
\label{eq:contrastiveloss}
\end{equation}
where $\tau$ assumes the role of a temperature parameter, and it is worth noting that all embeddings undergo $\ell_2$-normalization.

The query and key are two distinct augmentations of an identical sentence. The negative bank, denoted as $N$, comprises instances that act as negatives for each positive pair. Specifically, it can be defined as encompassing the remaining texts within the present minibatch \cite{chen2020simple}.

The logarithmic likelihood function formulated in Eq (\ref{eq:contrastiveloss}) operates within the bounds of a probability distribution constructed via the application of the \texttt{softmax} function to each input/query $q$. If we let $p_{z_i}$ denote the probability of match between the query and the feature $f_i \in F = N \cup {k}$, then the derivative of the loss concerning the query $q$, with respect to the gradient, can be expressed:
\begin{equation}
\begin{split}
\frac{\partial \mathcal{L}{q,k,{N}} }{\partial q} = - \frac{1}{\tau}& \left( (1 - p_k) \cdot k - \sum{n \in N} p_n \cdot n \right) \\
\textrm{ where } p_{f_i} &= \frac{\exp (q^T f_i / \tau) }{ \sum_{j \in Z} \exp (q^T f_j / \tau)}.
\label{eq:pi}
\end{split}
\end{equation}
$p_k$ and $p_n$ symbolize the likelihood of a match pertaining to the key and the negative feature, denoted as $f_i=k$ and $f_i=n$ respectively. It is discernible that the impacts of the positive and negative logits upon the loss function mirror those encountered in a ($K+1$)-way cross-entropy classification loss.

%%% 增加的内容
To further expand our understanding, let's consider the dimensional dynamics of the embeddings. Assume that each embedding vector $v$ (whether $q$, $k$, or $n$) is of dimension $d$. Then the normalization process can be described as:
\begin{equation}
\hat{v} = \frac{v}{| v |_2}
\end{equation}
where $| v |_2$ denotes the $L_2$ norm of vector $v$. This normalization ensures that the focus is on the direction of the vectors rather than their magnitude.

\subsubsection{Hard Negative Mixing for Contrastive Learning}

In contrastive learning, particularly within unsupervised learning, the role of hard negative samples is pivotal for effective model training. The standard practice of using batch negatives often falls short, as they may not offer a sufficient challenge or provide the necessary contrast with positive examples, as shown by recent studies \cite{hnm}. To address this, Hard Negative Mining has been introduced to enhance the learning process.

Hard Negative Mining involves deliberately selecting negative samples that are more challenging and closely resemble positive samples in the feature space. This method is more effective in teaching the model to distinguish between similar-looking but distinct categories, thereby improving the model's robustness and discriminative capabilities, as evidenced by recent advancements.

In response to the need for more effective hard negatives, we have developed an approach inspired by \cite{hnm} to further enhance the model's generalization capabilities. Our strategy innovatively applies the mixup technique to generate more relevant and challenging hard negative samples, thus refining the learning process.

As shown in Figure \ref{fig:hnm}, this approach begins by identifying a set of top $k$ vectors, denoted as $G=\{G_1, G_2,...,G_m\}$, which closely resemble a given input query $x_1$ from a previous minibatch. These vectors are then utilized as the foundation for crafting new hard negative samples. We employ a method of linear combination to produce $m$ novel hard negatives. These are subsequently incorporated into our existing negative sample bank (denoted as $N$ in equation \ref{eq:contrastiveloss}). The process of generating each new hard negative instance, $u_o$, is methodically formulated as follows:

\begin{equation}
u_o = \frac{\overline{u_o}}{||\overline{u_o}||_2}, \textrm{ where } \overline{u_o} = \alpha G_i + (1 - \alpha) G_j
\label{mixup}
\end{equation} 

Here, $u_o (where \ o \in \{1,...,m\})$ represents each new hard negative instance. The indices $(i, j)$ are randomly selected from the set of top $k$ vectors, and the coefficient $\alpha$ helps balance the contribution of each vector in the mixup.

Expanding further, we introduce an adaptive selection process for $\alpha$, which varies based on the similarity between the vectors $G_i$ and $G_j$. This approach ensures that the resulting hard negative is neither too similar nor too dissimilar to the query, maintaining an optimal level of difficulty. The adaptive selection is defined as:

\begin{equation}
\alpha = f(sim(G_i, G_j))
\end{equation} 

where $f$ is a function that maps the similarity score to an appropriate value for $\alpha$. The similarity function $sim(\cdot, \cdot)$ could be a cosine similarity.

This section outlines the Hard Negative Mixing strategy to optimize sentence representation learning. By crafting challenging negative samples, this approach enhances the model's semantic relationship recognition. The efficacy of this method is demonstrated through experiments detailed in the subsequent chapter, showcasing performance gains and comparisons with other methods.

The Appendix \ref{appendix} provides a comprehensive mathematical analysis of the Hard Negative Mixing strategy. It defines hard negatives, explains the generation of challenging negatives through linear combinations, and discusses the impact on the model's optimization and loss function adjustments. The analysis elucidates how this method enhances model performance by increasing sample distinctiveness and improving discriminative ability. It also addresses the role of regularization and learning rate adjustments in stabilizing and optimizing the learning process. This theoretical framework supports the experimental outcomes and offers guidance for applying the strategy effectively.

\section{Experiments}

\subsection{Experimental Settings}
We executed experiments across seven established semantic text similarity (STS) benchmarks, spanning STS 2012–2016 as delineated by Agirre et al. from 2012 to 2016, in conjunction with the STSBenchmark \cite{stsb} and SICKRelatedness \cite{sickr}. Our model's performance was assessed using the SentEval toolkit, and the Spearman’s correlation was chosen as the evaluation metric. Our methodology initiated by leveraging pre-existing BERT checkpoints. The [CLS] token embedding, extracted from the model's output, served as the sentence representation. Beyond semantic similarity evaluations, our model was subjected to seven distinct transfer learning tasks to gauge its capacity for generalization.

\subsection{Implementation Details}
In the methodology outlined by Gao, Yao, and Chen \cite{SimCSE}, we adopted an identical training dataset and protocol. This dataset encompasses a million sentences, sourced from Wikipedia. Utilizing a BERT model that's been finely optimized \cite{bert}, we derive an embedding for each of these sentences. Subsequent to this, we employ distinct dropout masks to produce augmented variants of the aforementioned sentence embeddings. All models are trained over a single epoch, with a batch size set at 64. Our computational implementation is founded on Python version 3.8.13, leveraging the capabilities of Pytorch version 1.12.1. All experimental tasks were executed on an NVIDIA GeForce RTX 3090 GPU equipped with 24G memory. 

\begin{table*}[t]
	\centering
        \renewcommand{\arraystretch}{1.2}
	\scalebox{0.8}{
	\begin{tabular}{lllllllll}
		\hline
		Model                           & STS12 & STS13 & STS14 & STS15 & STS16 & STS-B & SICK-R & \textbf{Avg.}   \\
		\hline
		\multicolumn{9}{c}{Unsupervised Models (Base)}                                                           \\
		\hline
		GloVe (avg.)                & 55.14 & 70.66 & 59.73 & 68.25 & 63.66 & 58.02 & 53.76  & 61.32  \\
		BERT (first-last avg.)      & 39.70 & 59.38 & 49.67 & 66.03 & 66.19 & 53.87 & 62.06  & 56.70  \\
		BERT-flow                   & 58.40 & 67.10 & 60.85 & 75.16 & 71.22 & 68.66 & 64.47  & 66.55  \\
		BERT-whitening              & 57.83 & 66.90 & 60.90 & 75.08 & 71.31 & 68.24 & 63.73  & 66.28  \\
		IS-BERT                     & 56.77 & 69.24 & 61.21 & 75.23 & 70.16 & 69.21 & 64.25  & 66.58  \\
		CT-BERT                     & 61.63 & 76.80 & 68.47 & 77.50 & 76.48 & 74.31 & 69.19  & 72.05  \\
		RoBERTa (first-last avg.)   & 40.88 & 58.74 & 49.07 & 65.63 & 61.48 & 58.55 & 61.63  & 56.57  \\
		RoBERTa-whitening           & 46.99 & 63.24 & 57.23 & 71.36 & 68.99 & 61.36 & 62.91  & 61.73  \\
		DeCLUTR-RoBERT              & 52.41 & 75.19 & 65.52 & 77.12 & 78.63 & 72.41 & 68.62  & 69.99  \\
		SIMCSE                     & 68.40 & 82.41 & 74.38 & 80.91 & 78.56 & 76.85 & 72.23  & 76.25  \\
        SIMCSE$_{(reproduce)}$                     & 70.82 & 82.24 & 73.25 & 81.38 & 77.06 & 77.24 & 71.16  & 76.16  \\
            LLaMA2-7B                & 50.66 & 73.32 & 62.76 & 67.00 & 70.98 & 63.28 & 67.40  & 65.06  \\
            LLaMA2-7B$_{(PromptEOL)}$ & 58.81 & 77.01 & 66.34 & 73.22 & 73.56 & 71.66 & 69.64  & 70.03  \\
            LLaMA2-7B$_{(Pretended\textunderscore CoT)}$ & 67.45 & 83.89 & 74.14 & 79.47 & 80.76 & 78.95 & 73.33  & 76.86  \\
            LLaMA2-7B$_{(Konwledge\textunderscore Enhancement)}$ & 65.60 & 82.82 & 74.48 & 80.75 & 80.13 & 80.34 & \textbf{75.89}  & 77.14  \\
        \textbf{HNCSE-PM(ours)}                     & \textbf{71.02} & \textbf{83.92} & \textbf{75.52} & \textbf{82.93} & \textbf{81.03} & \textbf{81.45} & 72.76  & \textbf{78.38}  \\
        \textbf{HNCSE-HNM(ours)}        & \textbf{69.76} & \textbf{83.97} & \textbf{75.52} & \textbf{83.21} & \textbf{81.63} & \textbf{81.85} & 72.87 & \textbf{78.27} \\
		\hline
		\multicolumn{9}{c}{Unsupervised Models (Large)}                                                   \\
		\hline
		SIMCSE                & 70.88 & 84.16 & 76.43 & 84.50 & 79.76 & 79.26 & 73.88  & 78.11  \\
            SIMCSE$_{(reproduce)}$                & 71.02 & 83.52 & 76.06 & 83.83 & 78.95 & 79.26 & 72.24  & 77.84  \\
 
        \textbf{HNCSE-PM(ours)}                      & \textbf{72.94} & \textbf{84.67} & \textbf{77.24} & 83.97 & 79.53 & \textbf{80.78} & \textbf{74.79}  & \textbf{79.13}  \\
        \textbf{HNCSE-HNM(ours)}                      & \textbf{72.75} & \textbf{84.54} & \textbf{77.36} & \textbf{84.58} & \textbf{79.92} & \textbf{80.60} & \textbf{74.64}  & \textbf{79.20}  \\
		\hline
	\end{tabular}
	}
	\caption{Spearman’s correlation scores across seven STS benchmarks for various models. HNCSE-PM refers to Positive Mixing method. HNCSE-HNM refers to Hard Negative Mixing method.}
	\label{table1}
\end{table*}

\begin{table*}[t]
	\centering
        \renewcommand{\arraystretch}{1.3}
	\scalebox{0.8}{
	\begin{tabular}{lllllllll}
		\hline
        Model & MR & CR & SUBJ & MPQA & SST & TREC & MRPC & \textbf{Avg.}   \\
        \hline
		\multicolumn{9}{c}{Unsupervised Models(Base)}  \\
		\hline
		GloVe (avg.) & 77.25 & 78.30 & 91.17 & 87.85 & 80.18 & 83.00 & 72.87 & 81.52  \\
		BERT & 78.66 & 86.25 & 94.37 & 88.66 & 84.40 & 92.80 & 69.54 & 84.94  \\
		IS-BERT & 81.09 & 87.18 & 94.96 & 88.75 & 85.96 & 88.64 & 74.24 & 85.83  \\

  SimCSE-RoBERTa & 81.04 & 87.74 & 93.28 & 86.94 & 86.60 & 84.60 & 73.68 & 84.84  \\

		\textbf{HNCSE-PM(ours)} & \textbf{81.94} & 86.99 & \textbf{95.20} & \textbf{89.77} & \textbf{86.81} & 85.31 & \textbf{75.49} & \textbf{85.93}  \\

    \textbf{HNCSE-HNM(ours)} & \textbf{81.64} & 86.84 & \textbf{95.11} & \textbf{89.66} & \textbf{86.81} & 83.90 & \textbf{75.91} & \textbf{85.70}  \\
		\hline
		\multicolumn{9}{c}{Unsupervised Models(Large)}                                                           \\
		\hline

		SimCSE-RoBERTa & 82.74 & 87.87 & 93.66 & 88.22 & 88.58 & 92.00 & 69.68 & 86.11   \\

		\textbf{HNCSE-PM(ours)} & \textbf{84.91} & \textbf{89.32} & \textbf{95.00} & \textbf{90.02} & \textbf{89.68} & 86.85 & \textbf{75.49} 	& \textbf{87.32}  \\

    \textbf{HNCSE-HNM(ours)} & \textbf{85.89} & \textbf{90.57} & \textbf{96.06} & \textbf{89.91} & \textbf{89.91} & 85.91 & \textbf{76.47} & \textbf{87.82}
    \\
		\hline
	\end{tabular}
	}
	\caption{Results on the Transfer Task datasets.}
        \label{table2}
\end{table*}

\subsection{Main Results}
\subsubsection{STS Task} 

In the study by \cite{mixcse}, we benchmarked the newly introduced HNCSE against a range of conventional unsupervised approaches and the prevailing state-of-the-art contrastive learning technique for the text semantic similarity (STS) task. This comparison encompasses methods such as average GloVe embeddings \cite{glove}, mean embeddings from BERT or RoBERTa, BERT-flow, BERT-whitening, ISBERT \cite{ISBERT}, DeCLUTR \cite{DeCLUTR}, CT-BERT \cite{CT-BERT}, and SimCSE \cite{SimCSE}.

As delineated in Table \ref{table1}, the peak performances achieved by our two variant models of HNCSE are 78.38\% and 78.27\% respectively. This clearly surpasses the unsupervised SimCSE employing a BERT-base architecture. Specifically, HNCSE exhibits superior performance to SimCSE across STS2012, STS2013, STS2014, STS2015, STS2016, SICK-R and STS-B benchmarks. Moreover, the more robust iterations of our HNCSE, leveraging the BERT-large architecture, consistently outshine the corresponding large model of SimCSE on the majority of STS tasks.

\subsubsection{Large language models are used for the STS Task} 
The growing interest in open-source large language models (LLMs) like LLaMA has highlighted their use in unsupervised sentence similarity tasks (STS). However, studies \cite{llm1,llm2} have shown that these models' performance in STS has not been up to par. The latest results for LLaMA2-7B and its three prompt engineering strategies—PromptEOL, Pretended Chain of Thought (CoT), and Knowledge Enhancement—have demonstrated a shortfall when compared to the methods introduced by HNCSE.

The architectures and training objectives of LLaMA2-7B and HNCSE have notably diverged. LLaMA2-7B, an autoregressive model, has been optimized for text generation, while HNCSE has been trained using contrastive learning for sentence-level representation. This divergence may have reduced the efficacy of LLaMA2-7B in sentence similarity tasks compared to HNCSE. Additionally, prompt engineering for LLaMA2-7B has faced limitations: (1) it heavily relies on carefully crafted templates requiring extensive optimization across models and tasks, increasing complexity in research and application; (2) these methods, designed primarily for generative models, show limited effectiveness for discriminative models, limiting their applicability; (3) as prompts become longer and more complex, the computational resource demand escalates, especially with large-scale datasets, highlighting resource consumption. In contrast, HNCSE's contrastive learning has eliminated the need for complex prompt designs, ensuring stable application across various models and tasks, and has shown improved efficiency in resource consumption.

\subsubsection{Transfer Task}

In the comprehensive evaluation of HNCSE's performance using the seven transfer tasks from SentEval \cite{tr}, the results delineated in Table \ref{table2} are particularly revealing. Notably, both the Base and Large configurations of HNCSE surpass SimCSE in six out of the seven benchmark datasets within the TR task. Importantly, a remarkable enhancement in performance is distinctly observed for the MRPC dataset.

\subsubsection{Ablation Study}
To assess model components' impact, we run ablation studies using two model variants, HNCSE-PM$_{single}$ and HNCSE-HNM$_{single}$, with identical setups and hyperparameters as the main experiment.

\textbf{HNCSE-PM$_{single}$} 
This model is a variant of HNCSE-PM, which means that in the Equation \ref{pos1} of Positive Mixing, we only optimize the case where $d_1$ is greater than $d_2$, and the case where $d_1$ is less than $d_2$ is not considered for optimization.

\textbf{HNCSE-HNM$_{single}$}
This model is a variant of HNCSE-HNM, which refers to the operation of directly removing Mixup in the Positive Mixing method, that is, directly replacing $u_o$ with $G_i$ in Equation \ref{mixup}.

Ablation results are shown in Table \ref{abl}. 
The experimental results demonstrate that the outcomes of the two HNCSE$_{single}$ methods are, on average, lower than the two HNCSE methods on both the STS task and the TR task. Therefore, the following conclusions can be drawn:

(1) It is essential to optimize the positive samples by leveraging the hardest negative samples, which helps to prevent these samples from becoming over-hard negative samples.
(2) The application of pairwise Mixup on hard negative samples is necessary, as it can capture beneficial aspects between different hard negative samples for the positive samples, thereby constructing more challenging negative samples.

\begin{table}[t]
\centering
\renewcommand{\arraystretch}{1.3}
\begin{tabular}{lcc}
\hline
Model & STS(Avg) & TR(Avg) \\
\hline
HNCSE-PM-BERT$_{base}$ & \textbf{78.38} & \textbf{85.93} \\
HNCSE-HNM-BERT$_{base}$ & \textbf{78.27} & \textbf{85.70} \\
HNCSE-PM$_{single}$-BERT$_{base}$ & 76.43 & 85.06 \\
HNCSE-HNM$_{single}$-BERT$_{base}$ & 76.64 & 84.89 \\
\hline

\end{tabular}
\vspace{0.2cm} % 调整标题与表格之间的距离
\caption{Results of the Ablation Study.}
\label{abl}
\end{table}

\subsubsection{Analysis of Batch Size and Max Sequence Length}

\begin{figure}[t!]
\centering
\subfigure[Impact of batch size]{\label{Fig-3a}
            \begin{tikzpicture}[font=\Large, scale=0.60]
                \begin{axis}[
                    legend cell align={left},
                    legend style={nodes={scale=1.0, transform shape}},
                    xlabel={Batch Size},
                    xtick pos=left,
                    tick label style={font=\large},
                    ylabel style={font=\large},
                    %xmode=log,
                    %log ticks with fixed point,
                    ylabel={Avg, Spearman},
                    xtick={1, 2, 3, 4, 5},
                    xticklabels={$16$, $32$, $64$, $128$, $256$},
                    ytick={73, 74, 75, 76, 77,78},
                    yticklabels={$73$, $74$, $75$, $76$, $77$,$78$},
                    legend pos=south east,
                    ymajorgrids=true,
                    grid style=dashed
                ]
                \addplot[
                    color=purple,
                    dotted,
                    mark options={solid},
                    mark=diamond*,
                    line width=1.5pt,
                    mark size=2pt
                    ]
                    coordinates {
                    (1, 73.96)
                    (2, 76.54)
                    (3, 78.37)
                    (4, 76.23)
                    (5, 75.76)

                    };
                    \addlegendentry{HNCSE-PM}
                % \addplot[
                %     color=purple,
                %     dotted,
                %     mark options={solid},
                %     mark=diamond*,
                %     line width=1.5pt,
                %     mark size=2pt
                %     ]
                %     coordinates {

                %     (1, 74.14)
                %     (2, 76.44)
                %     (3, 77.30)
                %     (4, 76.14)
                %     (5, 75.14)

                %     };
                %     \addlegendentry{HNCSE-2}
                \addplot[
                    color=blue,
                    dotted,
                    mark options={solid},
                    mark=*,
                    line width=1.5pt,
                    mark size=2pt
                    ]
                    coordinates {
                    (1, 73.24)
                    (2, 76.97)
                    (3, 78.27)
                    (4, 76.45)
                    (5, 76.06)

                    };
                    \addlegendentry{HNCSE-HNM}
                \end{axis}
                \end{tikzpicture}
            %\label{fig:flops_intro}
    }
    \subfigure[Impact of max sequence length]{\label{Fig-3b}
            \begin{tikzpicture}[font=\Large,scale=0.59]
                \begin{axis}[
                    legend cell align={left},
                    legend style={nodes={scale=1.0, transform shape}},
                    xlabel={Max Seq Length},
                    xtick pos=left,
                    tick label style={font=\large},
                    ylabel style={font=\large},
                    %title style={font=\large},
                    %xmode=log,
                    %log ticks with fixed point,
                    ylabel={ },
                    xtick={1, 2, 3, 4},
                    xticklabels={$16$, $32$, $64$, $128$, $256$},
                    ytick={74, 75, 76, 77, 78},
                    yticklabels={$74$, $75$,$76$,$77$,$78$},
                    legend pos=south east,
                    ymajorgrids=true,
                    grid style=dashed
                ]
                \addplot[
                    color=purple,
                    dotted,
                    mark options={solid},
                    mark=diamond*,
                    line width=1.5pt,
                    mark size=2pt
                    ]
                    coordinates {
                    (1, 74.67)
                    (2, 78.37)
                    (3, 76.71)
                    (4, 76.21)
                    };
                    \addlegendentry{HNCSE-PM}
               \addplot[
                    color=blue,
                    dotted,
                    mark options={solid},
                    mark=diamond*,
                    line width=1.5pt,
                    mark size=2pt
                    ]
                    coordinates {
                    (1, 75.67)
                    (2, 78.27)
                    (3, 76.95)
                    (4, 76.14)
                    };
                    \addlegendentry{HNCSE-HNM}
                \end{axis}
                \end{tikzpicture}
            %\label{fig:flops_intro}
    }
    
    % \subfigure[Result of selecting m]{\label{Fig-3c}
    %         \begin{tikzpicture}[font=\Large,scale=0.45]
    %             \begin{axis}[
    %                 legend cell align={left},
    %                 legend style={nodes={scale=1.0, transform shape}},
    %                 % title={MRR vs. \#modified Hous. $m$},
    %                 xlabel={m},
    %                 xtick pos=left,
    %                 tick label style={font=\large},
    %                 ylabel style={font=\large},
    %                 %title style={font=\large},
    %                 %xmode=log,
    %                 %log ticks with fixed point,
    %                 ylabel={Avg. Spearman},
    %                 %xmin=0.005, xmax=100.0,
    %                 % ymin=0.467,
    %                 xtick={0, 1, 2, 3, 4},
    %                 xticklabels={$0$, $1$, $2$, $3$, $4$},
    %                 ytick={75, 75.5, 76, 76.5, 77, 77.5},
    %                 yticklabels={$75$, $75.5$,$76$,$76.5$,$77$,$77.5$},
    %                 legend pos=south west,
    %                 ymajorgrids=true,
    %                 grid style=dashed
    %             ]
    %             \addplot[
    %                 color=blue,
    %                 dotted,
    %                 mark options={solid},
    %                 mark=diamond*,
    %                 line width=1.5pt,
    %                 mark size=2pt
    %                 ]
    %                 coordinates {
    %                 (0, 76.25)
    %                 (1, 77.14)
    %                 (2, 76.70)
    %                 (3, 76.43)
    %                 (4, 75.19)
    %                 };
    %                 \addlegendentry{HNCSE-3}
    %             \end{axis}
    %             \end{tikzpicture}
    %         %\label{fig:flops_intro}
    % }
    \vspace{-1mm}
    % \caption{(a) shows the influence of the hyperparameter batch size in the three methods of HNCSE. (b) illustrates the influence of the max sequence length in the HNCSE-HNM. (c) illustrates the influence of the number of additional embeddings (denoted as m) in the HNCSE-method3.}
     \caption{The influence of different hyperparameters on HNCSE.}
    \vspace{-2mm}

\end{figure}
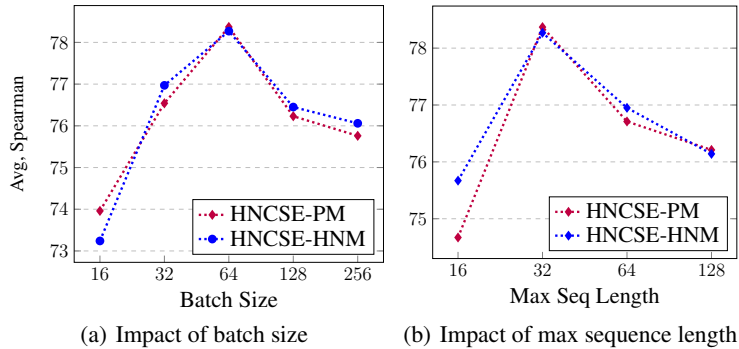

In our HNCSE model analysis, we've assessed the influence of batch size and max sequence length on model performance. Figure \ref{Fig-3a} demonstrates that optimal performance for both HNCSE-PM and HNCSE-HNM has been achieved with a batch size of 64, regardless of the number of hard negatives. The plateau in performance with larger batch sizes suggests that beyond a certain point, increased size no longer provides additional valuable learning information, underscoring the importance of hard negatives in mini-batches. Figure \ref{Fig-3b} shows the highest model value at a maximum sequence length of 32. Yet, model performance does not improve with increased sequence length. This may be due to longer sequences introducing greater ambiguities or complexities in semantic relationships, making it more challenging for the model to disambiguate and accurately capture nuances, thus reducing performance.

\section{Conclusion and Limitation}
In this work, we introduce a novel framework, HNCSE, specifically designed to address the multifaceted challenges posed by the incorporation of hard negative samples in sentence representation learning. Unlike conventional approaches, HNCSE innovatively integrates identified hard negative samples to optimize positive samples and construct even harder negative samples, thereby extending the well-established SimCSE methodology. This approach facilitates a deeper and more nuanced understanding of negative samples. Empirical evaluations of HNCSE on semantic textual similarity datasets and transfer task datasets demonstrate its superiority, indicating that unsupervised learning of sentence representations can make promising advancements to existing research.

Despite its strong performance, we have identified certain limitations associated with HNCSE. Firstly, HNCSE relies on large amounts of unannotated text data. If this data contains substantial noise or irrelevant information, it may adversely affect the quality of the model's embeddings. Secondly, HNCSE lacks explicit contextual information during training, potentially impeding its ability to capture long-distance dependencies. These issues present challenges for future research.

% \section*{References}

%\bibliographystyle{unsrtnat}
%\bibliography{Ref}

%%%%%%%%%%%%%%%%%%%%%%%%%%%%%%%%%%%%%%%%%%%%%%%%%%%%%%%%%%%%

\appendix

% \section{Appendix / supplemental material}
\section{Theoretical Analysis of Hard Negative Mixing}
\label{appendix}
Given a dataset of sentences $S = \{s_1, s_2, \ldots, s_n\}$, with a sentence embedding function $f: S \rightarrow \mathbb{R}^d$ mapping sentences to a $d$-dimensional vector space, \textbf{Hard Negative Mixing (HNM)} aims to refine the representation learning process by leveraging challenging negative samples.

\subsection{Definition of Hard Negatives}
Hard negatives are crucial for the effective training of models, especially in tasks that require fine-grained discrimination between semantically close but distinct sentences. A hard negative for a given sentence $s_i$ is defined as a sentence $s_j$ that, while semantically different, lies close in the embedding space. This closeness poses a greater challenge for the model to accurately discriminate:

\begin{equation}
HN(s_i) = \{s_j | sim(f(s_i), f(s_j)) > \theta, \forall s_j \in S \setminus \{s_i\}\}
\end{equation}

where $HN(s_i)$ denotes the set of hard negatives for the sentence $s_i$, $sim(\cdot)$ is a similarity measure (cosine similarity), $f(\cdot)$ represents the sentence embedding function mapping sentences to points in a high-dimensional vector space, $S$ is the set of all sentences, and $\theta$ is a predefined threshold indicating the minimum similarity for a sentence to be considered a hard negative.

\subsection{Mixing Strategy}
Building upon the foundational concept of hard negatives, as delineated in the preceding section, we introduce a novel approach to further enrich the training dataset and amplify the model's discriminative power. By strategically leveraging the defined hard negatives, we propose a method to generate synthetic samples that embody the nuances and complexities encountered in real-world scenarios. This method, termed Hard Negative Mixing (HNM), involves the innovative creation of synthetic hard negatives through the blending of embeddings from identified hard negative pairs. This strategy not only augments the diversity of training data but also introduces a heightened level of challenge, compelling the model to refine its understanding and representation of semantic distinctions.

To enhance the diversity and challenge of the training process, we employ a strategy of generating synthetic hard negatives. This is achieved by taking a linear combination of two existing hard negatives, $hn_i$ and $hn_j$, associated with the same anchor sentence:

\begin{equation}
hn_{mix} = \lambda hn_i + (1 - \lambda) hn_j
\end{equation}

where $\lambda$ is a mixing coefficient that lies in the range $[0,1]$. This coefficient determines the relative contribution of each hard negative to the synthetic sample. By adjusting $\lambda$, we can control the difficulty and variability of the generated hard negatives, enriching the training data and encouraging the model to learn more discriminative features.

\subsection{Optimization Objective}
The conceptual groundwork laid by the identification of hard negatives and the subsequent generation of synthetic hard negatives through our Hard Negative Mixing strategy paves the way for a refined training objective. This advanced stage of model training seeks not only to utilize these synthetic samples effectively but also to optimize the model's sensitivity towards differentiating between semantically similar and dissimilar sentences. By intricately balancing the influences of positive samples and synthetic hard negatives, we aim to craft an optimization objective that embodies the complexity of natural language, enhancing the model's ability to discern subtle semantic nuances.

The primary goal of incorporating HNM into the training process is to optimize the model's ability to distinguish between positive and negative samples. The optimization objective is formulated to minimize the distance between positive pairs while maximizing the distance between the anchor sentence and its synthetic hard negatives. This is quantified by the following loss function:

\begin{equation}
\begin{split}
\min_{f} \sum_{i=1}^{n} -\log \left( \frac{\exp(\text{sim}(f(s_i), f(s_i^+)) / \tau)}{\exp(\text{sim}(f(s_i), f(s_i^+)) / \tau) + \sum_{hn_{mix} \in HNM(s_i)} \exp(\text{sim}(f(s_i), f(hn_{mix})) / \tau)} \right)
\end{split}
\end{equation}

In this equation, $f(\cdot)$ denotes the embedding function mapping sentences to a high-dimensional vector space. The term $\text{sim}(\cdot)$ represents a similarity measure (e.g., cosine similarity) between two embeddings. The variable $s_i^+$ signifies a positive sample that is semantically similar to the sentence $s_i$, and $hn_{mix}$ represents a synthetic hard negative sample generated for $s_i$ through HNM. The parameter $\tau$, known as the temperature, scales the similarity scores to modulate the softmax function's concentration level, impacting the overall differentiation sensitivity.

The numerator of the fraction within the logarithm accentuates the affinity between $s_i$ and its positive counterpart $s_i^+$, compelling the model to cluster semantically similar sentences closer in the embedding space. Conversely, the denominator aggregates the exponential similarities between $s_i$ and both its positive pair and the set of synthetic hard negatives, promoting a spatial distinction between $s_i$ and the challenging negatives generated via HNM. Through this loss function, the optimization objective delicately orchestrates a balance between attraction and repulsion forces in the embedding space, fostering a nuanced understanding of semantic relationships critical for the advancement of natural language processing capabilities.

%%%%%%%%%%%
\subsection{Optimization Objective}

In the realm of representation learning, particularly within the framework of HNM, the essence of refining the model's capacity to distinguish between semantically proximate yet distinct sentences is encapsulated in the formulation of a meticulously crafted optimization objective. This advanced stage of model refinement leverages both the synthetic hard negatives generated through HNM and the positive samples to formulate a loss function that embodies the intricate dynamics of natural language understanding. The overarching goal is to optimize the model in such a way that it achieves a heightened sensitivity to the subtle semantic nuances that differentiate positive pairs from their hard negative counterparts.

To articulate this concept more concretely, we introduce a loss function designed to minimize the distance between positive sentence pairs while simultaneously maximizing the distance between each sentence and its associated synthetic hard negatives. This is achieved through a contrastive learning approach, where the model is encouraged to align closely with positive samples in the embedding space and diverge from the synthetic hard negatives. The loss function is formulated as follows:

% \begin{equation}
% \begin{split}
% % \min_{f} & \sum_{i=1}^{n} -\log \left( \frac{\exp(\text{sim}(f(s_i), f(s_i^+)) / \tau)}{\exp(\text{sim}(f(s_i), f(s_i^+)) / \tau) + } \right. \\
% % & \left. \sum_{hn_{mix} \in HNM(s_i)} \exp(\text{sim}(f(s_i), f(hn_{mix})) / \tau) \right)

% \min_{f} \sum_{i=1}^{n} -\log \left( \frac{\exp(\text{sim}(f(s_i), f(s_i^+)) / \tau)}{\exp(\text{sim}(f(s_i), f(s_i^+)) / \tau) + \sum_{hn_{mix} \in HNM(s_i)} \exp(\text{sim}(f(s_i), f(hn_{mix})) / \tau)} \right)

% \end{split}
% \end{equation}

\begin{equation}
\min_{f} \sum_{i=1}^{n} -\log \left( \frac{\exp(\text{sim}(f(s_i), f(s_i^+)) / \tau)}{\exp(\text{sim}(f(s_i), f(s_i^+)) / \tau) + \sum_{hn_{mix} \in HNM(s_i)} \exp(\text{sim}(f(s_i), f(hn_{mix})) / \tau)} \right)
\end{equation}

In this equation, $f(\cdot)$ denotes the embedding function that maps sentences to a $d$-dimensional vector space. The function $sim(\cdot)$ represents a measure of similarity (e.g., cosine similarity) between the embeddings of two sentences. The term $s_i^+$ refers to a positive sample that is semantically similar to $s_i$, and $hn_{mix}$ denotes a synthetic hard negative sample generated through the HNM process for the sentence $s_i$. The parameter $\tau$, known as the temperature, serves to scale the similarity scores, controlling the sharpness of the softmax distribution utilized in the denominator.

The numerator of the fraction within the logarithm focuses on the similarity between the sentence $s_i$ and its positive counterpart $s_i^+$, encouraging the model to learn embeddings that bring these pairs closer together in the embedding space. Conversely, the denominator aggregates the similarities between $s_i$ and its synthetic hard negatives $hn_{mix}$, alongside the similarity to the positive pair, thereby promoting the model to differentiate $s_i$ from its hard negatives effectively.

\subsection{Analyzing Impact on Embedding Space}
Following the in-depth exploration in the "Optimization Objective" section, where we discussed optimizing the model to distinguish between positive samples and synthetic hard negatives through a carefully designed loss function, this process not only enhances the model's discriminative capabilities but also suggests a significant impact on the structure of the embedding space. Therefore, our subsequent analysis will focus on the specific effects of the HNMstrategy on the embedding space, particularly on how it alters the average pairwise distance between embeddings, thereby facilitating a more distinct separation of semantic meanings.

The structural dynamics of the embedding space are crucial for the effective representation of semantic relationships among sentences. The introduction of mixed hard negatives via HNM alters this structure, potentially leading to a more separable space where similar and dissimilar sentences are better distinguished. This transformation can be quantified by measuring the average pairwise distance among embeddings before and after the application of HNM:

\begin{equation}
\begin{split}
\Delta = & \frac{1}{|S|^2} \sum_{s_i, s_j \in S} \|f(s_i) - f(s_j)\|_2 
- \frac{1}{|S|^2} \sum_{s_i, s_j \in S'} \|f'(s_i) - f'(s_j)\|_2
\end{split}
\end{equation}

% where $\Delta$ represents the change in average pairwise distance, indicating how the embedding space has expanded or contracted as a result of HNM. A larger $\Delta$ suggests a more discriminative embedding space, conducive to improved model performance.

In this equation, $S$ denotes the original set of sentences, and $S'$ represents the set of sentences after the application of HNM. The function $f$ maps sentences to their embeddings in the $d$-dimensional vector space, while $f'$ denotes the updated mapping post-HNM application. The norm $|\cdot|_2$ measures the Euclidean distance between two embeddings, providing a quantifiable measure of their dissimilarity.

The term $\Delta$ embodies the net change in the average pairwise distance among all sentence embeddings within the dataset, serving as an indicator of how the embedding space has evolved in response to HNM. A positive value of $\Delta$ suggests that, on average, embeddings have become more dispersed, implying that HNM has succeeded in expanding the embedding space. Such an expansion is indicative of a model that is better attuned to differentiating between semantically similar and dissimilar sentences, potentially leading to improvements in tasks requiring fine-grained semantic discrimination.

Conversely, a negative value of $\Delta$ would suggest a contraction of the embedding space, which could have implications for the model's ability to distinguish between closely related semantic concepts. However, the goal of HNM is not merely to expand the embedding space indiscriminately but to do so in a manner that enhances the model's discriminative capacity by fostering a more structured and separable embedding landscape.

\subsection{Enhancement of Model Discriminability}
Following the detailed exploration of HNMstrategies and their profound impact on the embedding space, our focus shifts towards quantitatively assessing the improvements in model discriminability. The synthetic hard negatives generated through HNM pose refined challenges, necessitating an advanced model capability to distinguish between semantically similar yet distinct sentences. This subsequent section delves into the empirical evaluation of model performance enhancements, highlighting the pivotal role of HNM in advancing the nuanced discernment capabilities of sentence embeddings.

One of the primary goals of HNM is to enhance the model's ability to discriminate between semantically similar and dissimilar sentences, which is pivotal for tasks such as semantic similarity measurement and classification. This enhancement can be theoretically represented by an increase in model accuracy, measured as follows:

\begin{equation}
\begin{split}
\mathcal{A}' - \mathcal{A} = \frac{1}{|T|} \sum_{(s_i, s_i^+, hn_{mix}) \in T} & \big[ \mathbb{I}[sim(f'(s_i), f'(s_i^+)) > 
 sim(f'(s_i), f'(hn_{mix}))] - \\
& \mathbb{I}[sim(f(s_i), f(s_i^+)) > 
sim(f(s_i), f(hn_{mix}))] \big]
\end{split}
\end{equation}

% where $\mathcal{A}$ and $\mathcal{A}'$ denote the accuracy of the model before and after applying HNM, respectively. This measure directly reflects the model's enhanced capability to differentiate between closely related sentences, thereby underlining the effectiveness of HNM in fostering more nuanced sentence representations.

In this formulation, $\mathcal{A}$ and $\mathcal{A}'$ represent the model's accuracy before and after the application of HNM, respectively. The set $T$ encompasses the triplets formed by an anchor sentence $s_i$, its positive counterpart $s_i^+$, and the synthetic hard negative $hn_{mix}$, generated through HNM. The function $\mathbb{I}[\cdot]$ denotes the indicator function, yielding a value of 1 when the condition within the brackets is true, and 0 otherwise. The term $sim(f(s_i), f(s_j))$ measures the similarity between the embeddings of sentences $s_i$ and $s_j$, facilitated by the embedding function $f(\cdot)$, which is refined to $f'(\cdot)$ post-HNM application.

The accuracy measure thus reflects the differential impact of HNM on the model's performance, quantified through the comparison of similarity scores between positive and hard negative pairs, before and after the integration of HNM. This metric directly illuminates the enhanced discriminative capability of the model, as a result of the nuanced challenges introduced by synthetic hard negatives, fostering a deeper understanding and distinction between closely related sentences.

\subsection{Impact on Learning Dynamics}
After elucidating the role of HNMin enhancing model discriminability and its substantial effects on the embedding space, it becomes imperative to examine how these strategies influence the learning dynamics of the model. The adaptive challenges introduced by synthetic hard negatives necessitate a deeper understanding of the underlying adjustments in the model's optimization process. This analysis is critical for optimizing the training strategy and ensuring that the model effectively leverages the nuanced complexities introduced by HNM, thereby achieving a sophisticated balance between accuracy and generalization.

The introduction of HNM into the training process modifies the learning dynamics by influencing the gradient of the loss function with respect to the embeddings. This section delves into the mathematical underpinnings of how HNM alters the embedding updates during training, providing insights into the optimization trajectory and the embedding space's evolution.

\subsubsection{Gradient of Loss Function}

The loss function's gradient with respect to the embeddings is a critical factor in the model's learning process, dictating how the embeddings are updated in each iteration. Given the loss function $L$, the gradient with respect to the embedding of a sentence $s_i$ can be expressed as:

\begin{equation}
\frac{\partial L}{\partial f(s_i)} = -\frac{1}{\tau} \left( (1 - p_k) \cdot k - \sum_{n \in N} p_n \cdot n \right)
\end{equation}

where:
\begin{itemize}
    \item $f(s_i)$ denotes the embedding of sentence $s_i$.
    \item $\tau$ is the temperature parameter scaling the similarity scores.
    \item $p_k$ and $p_n$ represent the softmax probabilities of the positive and negative samples, respectively.
    \item $k$ is the embedding of the positive sample, and $n$ denotes the embeddings of the negative samples in the set $N$.
\end{itemize}

This formulation encapsulates the essence of contrastive learning, emphasizing the model's aim to pull positive pairs closer while pushing negative pairs apart in the embedding space.

HNM influences the learning dynamics by adjusting the distribution and characteristics of negative samples, thereby affecting the gradient of the loss function. The modification of the gradient can lead to more effective learning, especially in distinguishing closely related but distinct sentences. This is achieved by incorporating a mix of hard negatives, which are closer to the query sentence in the embedding space, thereby providing a stronger learning signal for the model.

\subsection{Balancing Diversity and Difficulty}
In the context of the profound adjustments introduced by HNMin the learning dynamics and its pivotal role in enhancing model discriminability, the strategic balance between the diversity and difficulty of hard negatives emerges as a critical consideration. This balance is instrumental in ensuring that the synthetic hard negatives generated through HNM not only challenge the model but also contribute to a comprehensive representation of the semantic landscape. Addressing this balance enables the optimization of the model's performance by fine-tuning the difficulty level of training samples to match the model's learning stage, thereby facilitating a more nuanced understanding of complex semantic relationships.

The efficacy of HNM is not merely a function of introducing hard negatives but also depends on balancing the diversity and difficulty of these negatives. This balance can be formalized as an optimization problem, aimed at maximizing the diversity among hard negatives while ensuring they remain challenging:

\begin{equation}
\max_{HN(s_i)} D(HN(s_i)) - \beta \sum_{hn \in HN(s_i)} sim(f(s_i), f(hn))
\end{equation}

% where $D(HN(s_i))$ quantifies the diversity among hard negatives for sentence $s_i$, and $\beta$ is a parameter that controls the trade-off between diversity and difficulty. An optimal balance ensures that the model is exposed to a wide range of challenging negatives, promoting a more comprehensive understanding of semantic nuances.

% By delving into these aspects of Hard Negative Mixing, we gain valuable insights into its capacity to enhance representation learning, particularly in natural language processing tasks where semantic precision is paramount.

Here, $HN(s_i)$ denotes the set of hard negatives for the sentence $s_i$, and $D(HN(s_i))$ quantifies the diversity within this set. The diversity metric is pivotal for ensuring that the model encounters a broad spectrum of semantic variations, thereby enhancing its generalizability. The term $sim(f(s_i), f(hn))$ measures the similarity between the embeddings of $s_i$ and its hard negatives $hn$, with $f(\cdot)$ representing the embedding function. The coefficient $\beta$ is a balancing parameter that modulates the emphasis between maximizing diversity and the aggregate similarity of the hard negatives, signifying the difficulty aspect.

This equation effectively captures the dual objective of optimizing the set of hard negatives by maximizing their diversity while controlling their difficulty, as measured by their similarity to the sentence $s_i$. The trade-off, facilitated by $\beta$, allows for a nuanced adjustment of the training process, tailoring the challenge level to optimally advance the model's discriminative capabilities.

\subsection{Optimization Objective with Regularization}
Following the exploration of the strategic balance between diversity and difficulty in HNMand its significant implications for model training dynamics, the introduction of a regularization term in the optimization objective becomes a logical progression. This approach seeks to address the potential for overfitting that may arise from the intensified training regimen implied by HNM. Incorporating regularization into the optimization framework not only ensures model robustness but also facilitates the generalization of learned embeddings across unseen data, thus safeguarding against the pitfalls of over-specialization to the training set.

Incorporating a regularization term, denoted as $R(f)$, into the optimization objective is a well-established method for enhancing the robustness of machine learning models. It penalizes complex models to prevent overfitting, thereby improving generalization:

\begin{equation}
\begin{aligned}
\min_{f} \Bigg[ & \sum_{i=1}^{n} -\log \Bigg( \frac{\exp(\text{sim}(f(s_i), f(s_i^+)) / \tau)}{\exp(\text{sim}(f(s_i), f(s_i^+)) / \tau) + \sum_{hn_{mix} \in HNM(s_i)} \exp(\text{sim}(f(s_i), f(hn_{mix})) / \tau)} \Bigg) \\
& + \lambda R(f) \Bigg]
\end{aligned}
\end{equation}

In this expression, $f(\cdot)$ represents the embedding function mapping sentences to their corresponding vectors in a $d$-dimensional space. The $sim(\cdot)$ function denotes a similarity measure (e.g., cosine similarity) between two sentence embeddings. The symbol $s_i^+$ refers to a positive sample associated with $s_i$, and $hn_{mix}$ represents the synthetic hard negatives generated through the HNM process for $s_i$. The parameter $\tau$, known as the temperature, is used to scale the similarity scores, influencing the sharpness of the softmax distribution in the denominator. The term $\lambda R(f)$ introduces the regularization component, where $\lambda$ is a coefficient that balances the relative importance of the regularization term within the overall optimization objective. The function $R(f)$ quantifies the complexity of the model, penalizing configurations that may lead to overfitting, thereby encouraging the model to adopt simpler, more generalizable forms.

This equation, therefore, encapsulates a dual objective: minimizing the contrastive loss to enhance discriminability and optimizing model complexity to prevent overfitting. By achieving this balance, the model is not only trained to distinguish effectively between semantically similar and dissimilar sentences but also ensured to generalize well to unseen data, embodying the essence of robust machine learning models.

\subsection{Temperature Scaling Analysis}
Following the intricate discussions on the optimization objectives, including the incorporation of regularization, and the strategic endeavors to balance diversity and difficulty within the HNMframework, the exploration of temperature scaling emerges as a crucial analytical component. This analysis is imperative for fine-tuning the model's sensitivity to distinctions between hard negatives and other sentence pairs. Temperature scaling, by adjusting the sharpness of the softmax distribution used in the model's loss function, plays a pivotal role in modulating the learning process. This nuanced adjustment allows for a more precise calibration of the model's response to the varying degrees of challenge presented by the hard negatives, thereby optimizing the effectiveness of the HNM strategy.

The temperature scaling parameter $\tau$ plays a crucial role in controlling the sharpness of the softmax function, directly impacting the model's learning dynamics:

% \begin{equation}
% \frac{\partial L}{\partial \tau} = -\frac{1}{\tau^2} \sum_{i=1}^{n} \Bigg( \frac{\exp(sim(f(s_i), f(s_i^+)) / \tau) \cdot sim(f(s_i), f(s_i^+))}{\exp(sim(f(s_i), f(s_i^+)) / \tau) + \sum_{hn_{mix} \in HNM(s_i)} \exp(sim(f(s_i), f(hn_{mix})) / \tau)} \Bigg).
% \end{equation}

To further enhance the readability of the long equation, we introduce intermediate variables to simplify its expression. This approach allows for a clearer understanding of each component of the equation. Let's define the following intermediate variables:

1. \(sim_{i}^{+} = sim(f(s_i), f(s_i^+))\), representing the similarity between the positive pair.

2. \(sim_{i, hn_{mix}} = sim(f(s_i), f(hn_{mix}))\), representing the similarity between the hard negative mix pair.

3. \(Z_i = \exp\left(\frac{sim_{i}^{+}}{\tau}\right) + \sum_{hn_{mix} \in HNM(s_i)} \exp\left(\frac{sim_{i, hn_{mix}}}{\tau}\right)\), representing the normalization factor.

With these definitions, we can rewrite the original equation as follows:

\begin{align}
\frac{\partial L}{\partial \tau} = & -\frac{1}{\tau^2} \sum_{i=1}^{n} \left( \frac{\exp\left(\frac{sim_{i}^{+}}{\tau}\right) \cdot sim_{i}^{+}}{Z_i} \right)
\label{eq:example}
\end{align}

This expression reveals how sensitivity to changes in $\tau$ affects the balance between positive and hard negative pairs in the loss gradient, guiding the optimization process.

\subsection{Embedding Space Normalization}
Building upon the insightful analyses of temperature scaling and its pivotal role in fine-tuning the model's discrimination capabilities, the focus now shifts toward the normalization of the embedding space. Embedding space normalization is a critical step in ensuring that the refined embeddings, influenced by HNM and temperature adjustments, maintain a consistent scale. This normalization facilitates a uniform interpretation of distances and similarities within the embedding space, which is essential for the effective application of HNM strategies. By normalizing the embeddings, we aim to enhance the model's generalization ability and ensure that the semantic nuances captured through HNM are robustly integrated into the model's learning framework.

When integrating HNM with embedding normalization to unit length, the process is adapted to accommodate the discriminative enhancements introduced by HNM. In the adjusted Equation:

\begin{equation}
f_{norm-HNM}(s_i) = \frac{f_{HNM}(s_i)}{\|f_{HNM}(s_i)\|_2}
\end{equation}

$f_{HNM}(s_i)$ is the embedding of sentence $s_i$ after applying HNM, reflecting the impact of challenging mixed negatives on the representation. The denominator $\|f_{HNM}(s_i)\|_2$ is the $L2$ norm of this embedding, ensuring normalization to unit length. This normalization focuses on the directional attributes of the embedding vectors in the high-dimensional space, crucial for capturing angular differences indicative of semantic relations. The result, $f_{norm-HNM}(s_i)$, is a normalized embedding that retains the rich, discriminative features imbued by HNM, facilitating a precise and nuanced separation of semantic content in the embedding space.

\subsection{Similarity Gradient Computation}
Transitioning from the essential task of embedding space normalization, which ensures uniformity and comparability of embeddings, we now advance to the intricate process of similarity gradient computation. This step is pivotal for understanding how the model updates its embeddings in response to the learning signal, especially within the HNM framework. The computation of similarity gradients is fundamental to the optimization process, as it elucidates the direction and magnitude of adjustments needed for the embeddings to accurately reflect semantic relationships. By calculating these gradients, we aim to further refine the model's ability to discern subtle semantic differences, leveraging the complexity and challenges introduced by synthetic hard negatives.

The gradient of the similarity function with respect to embeddings is essential for understanding how the model updates embeddings in response to the learning signal:

\begin{equation}
\begin{split}
\nabla_{f(s_i)} sim(f(s_i), f(s_j)) = & \frac{f(s_j)}{\|f(s_i)\|\|f(s_j)\|} 
 - \frac{sim(f(s_i), f(s_j))(f(s_i))}{\|f(s_i)\|^2}
\end{split}
\end{equation}

% This equation provides insights into the direction and magnitude of updates to the embedding of sentence $s_i$, driven by its relationship with sentence $s_j$.

In this expression, $\nabla_{f(s_i)} sim(f(s_i), f(s_j))$ represents the gradient of the similarity function between the embeddings of sentences $s_i$ and $s_j$ with respect to the embedding of sentence $s_i$. The function $f(\cdot)$ yields the embedding of a sentence, mapping it to a point in the embedding space. The term $sim(f(s_i), f(s_j))$ quantifies the similarity between the embeddings of $s_i$ and $s_j$, which is often calculated using cosine similarity or another relevant metric in the embedding space.

The gradient is composed of two primary components. The first component, $\frac{f(s_j)}{|f(s_i)||f(s_j)|}$, reflects the direction towards the embedding of $s_j$ normalized by the magnitudes of both embeddings, essentially pointing in the direction to increase similarity. The second component, $- \frac{sim(f(s_i), f(s_j))(f(s_i))}{|f(s_i)|^2}$, adjusts this direction by considering the current level of similarity and the magnitude of $f(s_i)$, thereby ensuring that the update is proportional to the need to increase or decrease similarity. This computation effectively guides the updates to embeddings during the learning process, ensuring that they evolve in a manner conducive to maximizing the discriminative capability of the model as influenced by the semantic relationships embodied in the training data.

\subsection{Analyzing Embedding Divergence}
Following the computation of similarity gradients, which provides crucial insights into the directional adjustments required for embeddings, we pivot towards the analysis of embedding divergence. This analysis is imperative as it quantitatively measures the extent to which the HNMstrategy influences the representational space of the embeddings. By evaluating the divergence between embeddings before and after the application of HNM, we can gauge the effectiveness of this strategy in enriching the model's semantic understanding. This step is not only logical but necessary for validating the impact of HNM on enhancing the model's discrimination capabilities, ensuring that the introduction of synthetic hard negatives leads to a more nuanced and robust representation of semantic relationships.

The divergence in embeddings, before and after applying HNM, sheds light on the method's effectiveness in altering the representational space:

\begin{equation}
div(f, f') = \sqrt{\sum_{i=1}^{n} \|f(s_i) - f'(s_i)\|_2^2}
\end{equation}

% This metric quantifies the extent of change in embeddings, indicating how HNM reshapes the embedding space to better capture semantic distinctions.
In this formulation, $div(f, f')$ quantifies the overall divergence between the embeddings produced by the model before ($f$) and after ($f'$) the application of HNM. The function $f(s_i)$ represents the embedding of sentence $s_i$ before the application of HNM, and $f'(s_i)$ denotes the embedding of the same sentence after applying HNM. The operation $|f(s_i) - f'(s_i)|_2$ computes the Euclidean distance between the original and updated embeddings for each sentence $s_i$ in the dataset, with the square root of the sum of these squared distances providing a holistic measure of how the embedding space has shifted as a result of HNM.

This divergence metric effectively encapsulates the extent of change in the embedding space, offering a concise measure of the impact of incorporating synthetic hard negatives through HNM on the model's semantic representation capabilities.

\subsection{Impact of Hard Negative Mixing on Similarity Distribution}
Building on our analysis of embedding divergence, which quantitatively demonstrates the transformative impact of HNMon the model's representational space, we next turn our attention to the broader implications of HNM on similarity distribution. This analysis is critical for understanding how HNM reshapes the landscape of semantic relationships within the embedding space, influencing the model's ability to discern and categorize these relationships accurately. By examining the shift in similarity distribution before and after the application of HNM, we seek to uncover the nuanced ways in which HNM enhances or modifies the model's perception of semantic similarity, thus providing valuable insights into the effectiveness and strategic value of incorporating hard negatives in the training process.

HNM fundamentally alters the similarity distribution between sentence pairs, an effect that can be measured using the Kullback-Leibler divergence:

\begin{equation}
D_{KL}(P_{sim} \| Q_{sim}) = 
\sum_{(s_i, s_j) \in S \times S} P_{sim}(s_i, s_j) \log \left( \frac{P_{sim}(s_i, s_j)}{Q_{sim}(s_i, s_j)} \right)
\end{equation}

In this equation, $D_{KL}(P_{sim} | Q_{sim})$ denotes the KL divergence between two probability distributions: $P_{sim}$, representing the distribution of similarities between sentence pairs before the application of HNM, and $Q_{sim}$, representing the distribution after applying HNM. The summation runs over all pairs of sentences $(s_i, s_j)$ within the set $S$, encompassing the entire dataset.

$P_{sim}(s_i, s_j)$ and $Q_{sim}(s_i, s_j)$ are the probabilities associated with the similarity scores between sentences $s_i$ and $s_j$ in the respective distributions before and after HNM is applied. The $\log$ function provides a measure of the relative difference between these probabilities, with the overall KL divergence summing these differences across all sentence pairs. This formulation effectively captures the divergence in similarity distributions due to HNM, providing a quantitative lens through which the impact of HNM on the embedding space's structure can be assessed.

\subsection{Enhanced Discriminative Feature Learning}
Building on the comprehensive analyses of embedding divergence and its implications for the representational space, we now turn our attention to the enhanced learning of discriminative features facilitated by HNM. This progression is both logical and necessary, as the previous analyses provide a foundation for understanding how HNM reshapes the embedding landscape. The subsequent enhancement in the model's ability to learn discriminative features is a direct consequence of these structural changes. By focusing on this aspect, we aim to elucidate the direct link between the strategic introduction of synthetic hard negatives and the model's improved capability to distinguish between semantically similar but distinct sentences. This step underscores the pivotal role of HNM in fostering a deeper, more nuanced comprehension of semantic relations, which is critical for advancing the state of the art in natural language processing tasks.

The ability of HNM to encourage the learning of more discriminative features is captured by the increased margin between positive and hard negative pairs:

\begin{equation}
\begin{split}
\Delta margin = \min_{s_i^+ \in P(s_i),\, hn_{mix} \in HNM(s_i)} \Big( & sim(f(s_i), f(s_i^+)) 
 - sim(f(s_i), f(hn_{mix})) \Big)
\end{split}
\end{equation}

In this formulation, $\Delta margin$ quantifies the increment in the separation margin between the embeddings of positive pairs ($s_i$, $s_i^+$) and the embeddings of sentence $s_i$ and its synthetic hard negatives $hn_{mix}$. The function $sim(f(s_i), f(s_j))$ calculates the similarity between the embeddings of sentences $s_i$ and $s_j$, with $f(\cdot)$ representing the embedding function. The set $P(s_i)$ contains positive samples that are semantically close to $s_i$, and $HNM(s_i)$ encompasses the synthetic hard negatives generated for $s_i$ through the HNM process.

The objective of this measure is to capture the extent to which the model has enhanced its ability to embed semantically similar sentences closer in the embedding space, while effectively distancing the hard negatives. A larger $\Delta margin$ signifies a more pronounced discriminative capability, highlighting the efficacy of HNM in enriching the feature learning process. This metric thus serves as a direct indicator of the model's improved proficiency in parsing and understanding the nuanced semantic relationships inherent in natural language.

\subsection{Learning Rate Adjustment Strategy}
After delving into the nuances of embedding divergence and its implications for the model's semantic representation, the progression towards refining the learning rate adjustment strategy becomes both logical and necessary. This refinement is pivotal in accommodating the enhanced discriminative feature learning facilitated by the HNMstrategy. By dynamically adjusting the learning rate in response to the presence of synthetic hard negatives, the model can more effectively integrate the complexity and challenge they introduce. This tailored approach to learning rate adjustment ensures that the model remains sensitive to subtle semantic distinctions, optimizing its performance across varying training dynamics. Thus, the strategic modulation of the learning rate emerges as an essential component of the optimization framework, enabling the model to fully leverage the benefits of HNM.

Adapting the learning rate in response to the presence of hard negatives provides a mechanism for fine-tuning the model's sensitivity to challenging samples:

\begin{equation}
\eta_{new} = \eta \cdot \gamma^{\mathbb{I}[hn_{mix} \in HNM(s_i)]}
\end{equation}

In this expression, $\eta_{new}$ represents the adjusted learning rate, while $\eta$ denotes the original learning rate prior to adjustment. The term $\gamma$ is a decay factor, with a value less than 1, which modulates the learning rate. The function $\mathbb{I}[\cdot]$ is an indicator function that evaluates to 1 if the condition within the brackets is true—specifically, if the synthetic hard negative $hn_{mix}$ is among the hard negatives generated for sentence $s_i$ through HNM—and 0 otherwise.

This formulation encapsulates a dynamic adjustment mechanism, wherein the learning rate is selectively scaled down in the presence of synthetic hard negatives. This scaling is designed to temper the learning step, acknowledging the heightened challenge posed by these negatives, and thereby ensuring that the model's updates are both deliberate and measured. The application of this strategy ensures that the learning process remains robust and responsive to the intricacies introduced by HNM, optimizing the model's trajectory towards achieving superior discriminative performance.
%%%%%%%%%%%%%%%%%%%%%%%%%%end

\subsection{Conclusion}
In conclusion, the theoretical exploration of Hard Negative Mixing (HNM) within this discourse elucidates its profound impact on the enhancement of representation learning, particularly in the domain of natural language processing. By rigorously defining hard negatives, devising a synthetic generation strategy, and optimizing the model through innovative loss functions, HNM has demonstrated significant promise in refining the discriminability and semantic understanding of models. The strategic incorporation of embedding space normalization, similarity gradient computation, and an adaptive learning rate adjustment strategy further optimizes the learning process, ensuring the model's sensitivity to nuanced semantic differences. The analysis of embedding divergence and the adjustment of similarity distributions underscore the effectiveness of HNM in creating a more nuanced and robust representational space. Ultimately, this comprehensive theoretical analysis highlights the paramount importance of HNM in advancing the state of the art in representation learning. Through meticulous examination and optimization, HNM emerges not merely as a technique but as a cornerstone in the pursuit of more discriminative, nuanced, and contextually aware models, underscoring the indispensability of theoretical analysis in pushing the boundaries of what is achievable in natural language processing and beyond.

% Optionally include supplemental material (complete proofs, additional experiments and plots) in appendix.
% All such materials \textbf{SHOULD be included in the main submission.}

%%%%%%%%%%%%%%%%%%%%%%%%%%%%%%%%%%%%%%%%%%%%%%%%%%%%%%%%%%%%

\end{document}